\documentclass[times,twocolumn,final]{elsarticle}

\usepackage{booktabs}
\usepackage{yjvci}
\usepackage{framed,multirow}

\usepackage{amssymb}
\usepackage{latexsym}

\usepackage{url}
\usepackage{xcolor}

\usepackage{hyperref}

\definecolor{newcolor}{rgb}{.8,.349,.1}

\journal{Journal of Visual Communication and Image Representation}

\begin{document}

\verso{Given-name Surname \textit{et~al.}}

\begin{frontmatter}

\title{Register assisted aggregation for Visual Place Recognition}

\author[1]{Xuan \snm{Yu}\fnref{fn1}}

\author[1]{Zhenyong \snm{Fu}\corref{cor1}}
\cortext[cor1]{Corresponding author: 
  e-mail: z.fu@njust.edu.cn;}
\fntext[fn1]{e-amil: xuan\_yu@njust.edu.cn;}


\address[1]{School of Computer Science and Engineering, Nanjing University of Science and Technology, Nanjing, 210094, China}

\received{-}
\finalform{-}
\accepted{-}
\availableonline{-}
\communicated{-}

\begin{abstract}
Visual Place Recognition (VPR) refers to the process of using computer vision to recognize the position of the current query image. Due to the significant changes in appearance caused by season, lighting, and time spans between query images and database images for retrieval, these differences increase the difficulty of place recognition. Previous methods often discarded useless features (such as sky, road, vehicles) while uncontrolled discarding features that help improve recognition accuracy (such as buildings, trees). To preserve these useful features, we propose a new feature aggregation method to address this issue. Specifically, in order to obtain global and local features that contain discriminative place information, we added some registers on top of the original image tokens to assist in model training. After reallocating attention weights, these registers were discarded. The experimental results show that these registers surprisingly separate unstable features from the original image representation and outperform state-of-the-art methods.
\end{abstract}


\end{frontmatter}


\section{Introduction}
Visual Place Recognition (VPR) aims to retrieve the most matching query image from a visual scene image database containing geographic location markers, so as to estimate the position information of the current query image. VPR has long been widely used in mobile robots \cite{xu2020probabilistic} and augmented reality \cite{middelberg2014scalable}, such as autonomous driving \cite{doan2019scalable}, image geolocation \cite{liu2019stochastic}, and 3D reconstruction \cite{liu2019lpd}. Its main challenges include changes in conditions (such as lighting, weather, and seasons), viewpoint changes, perceptual aliasing, and appearance changes over time.

The working principle of a VPR system is to represent a given query image as a compact descriptor, and then match it with a reference image database containing geographic location information. The traditional VPR method \cite{jegou2010aggregating, jegou2011aggregating, kim2015predicting} uses local aggregation descriptor vectors to retrieve the position of images. With the development of deep learning, convolutional neural networks (CNN) and transformer models \cite{vaswani2017attention} have shown excellent performance in computer vision tasks, including image classification, object detection, and semantic segmentation. Due to the self-attention mechanism of transformer models being able to establish associations between different places, and it can capture global and local relationships, as well as correlations between different regions in the image, thus effectively extracting important features in the image, many researchers have proposed using transformer models for VPR tasks, such as \cite{wang2022transvpr} and \cite{zhu2023r2former}, which have achieved great success.

\begin{figure*}[!t]
\centering
\includegraphics[scale=.6]{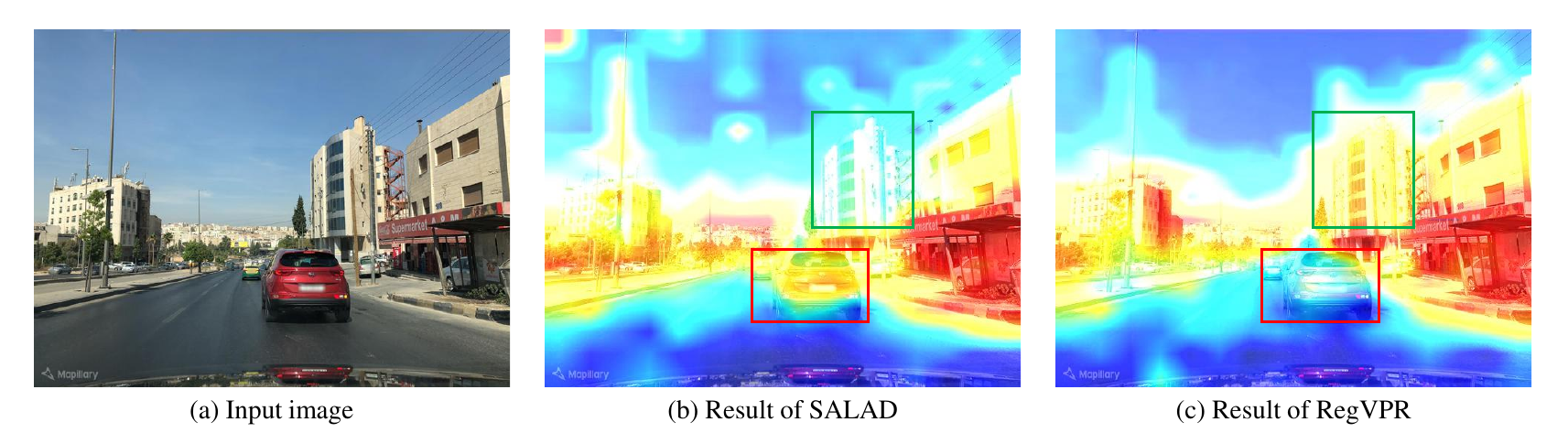}
\caption{\label{f1}Comparison of heatmap between SALAD model and our method. It can be seen intuitively that the SALAD model has discarded some of the building features (within the green box, which we hope to preserve), but has retained some of the vehicle features (within the red box, which we hope to discard).}
\end{figure*}

Despite the impressive performance of these methods, the features pre-trained with transformer often differ from the specific requirements of VPR tasks, making it difficult to fully utilize the performance of pre-trained models when directly applied to VPR tasks. These pre-trained models tend to aggregate unstable dynamic information (such as vehicles and pedestrians) into descriptors and tend to ignore some robust static discriminative information (such as buildings and plants), which is an undesirable phenomenon.

Recently, TransVPR \cite{wang2022transvpr}, SALAD \cite{izquierdo2023optimal}, and SelaVPR \cite{lu2024towards} have achieved excellent performance in many computer vision tasks using transformer models. The work of SALAD \cite{izquierdo2023optimal} follows the approach of NetVLAD \cite{arandjelovic2016netvlad}, quantifying local descriptors into a set of clusters. The difference is that the former uses optimal transmission algorithms to redefine features for cluster allocation and introduces a `Dustbin' mechanism to discard uninformed features, while the latter aggregates local descriptors by quantizing them into a set of clusters and storing the sum of residuals per cluster. However, such a `Dustbin' mechanism can effectively discard useless information (such as vehicles), but it also discards some robust feature representations (such as buildings), as shown in Fig. \ref{f1}. Inspired by \cite{darcet2023vision}, these discarded information are considered as outlier markers, which store global image information and typically appear in the background area of the image. The use of registers can effectively eliminate outliers in the image.

In this article, we propose a new method, which uses \textbf{Reg}isters to assist in removing irrelevant information from image representations in \textbf{VPR} tasks while preserving valid information, called \textbf{RegVPR}. Our method introduces registers during the feature aggregation process and uses a Transformer Encoder containing self-attention mechanism to reassign feature weights on the original image tokens and the local descriptor sequence after register concatenation. These registers can effectively capture these tokens containing a large amount of background information and discard them without compromising the quality of descriptor representation. We use pre-trained DINOv2 \cite{oquab2023dinov2} as our backbone and introduce some lightweight adapters to fine-tune the pre-trained backbone, thus enabling the pre-trained foundation model to seamlessly adapt to VPR tasks.

\begin{figure*}[!t]
\centering
\includegraphics[scale=.45]{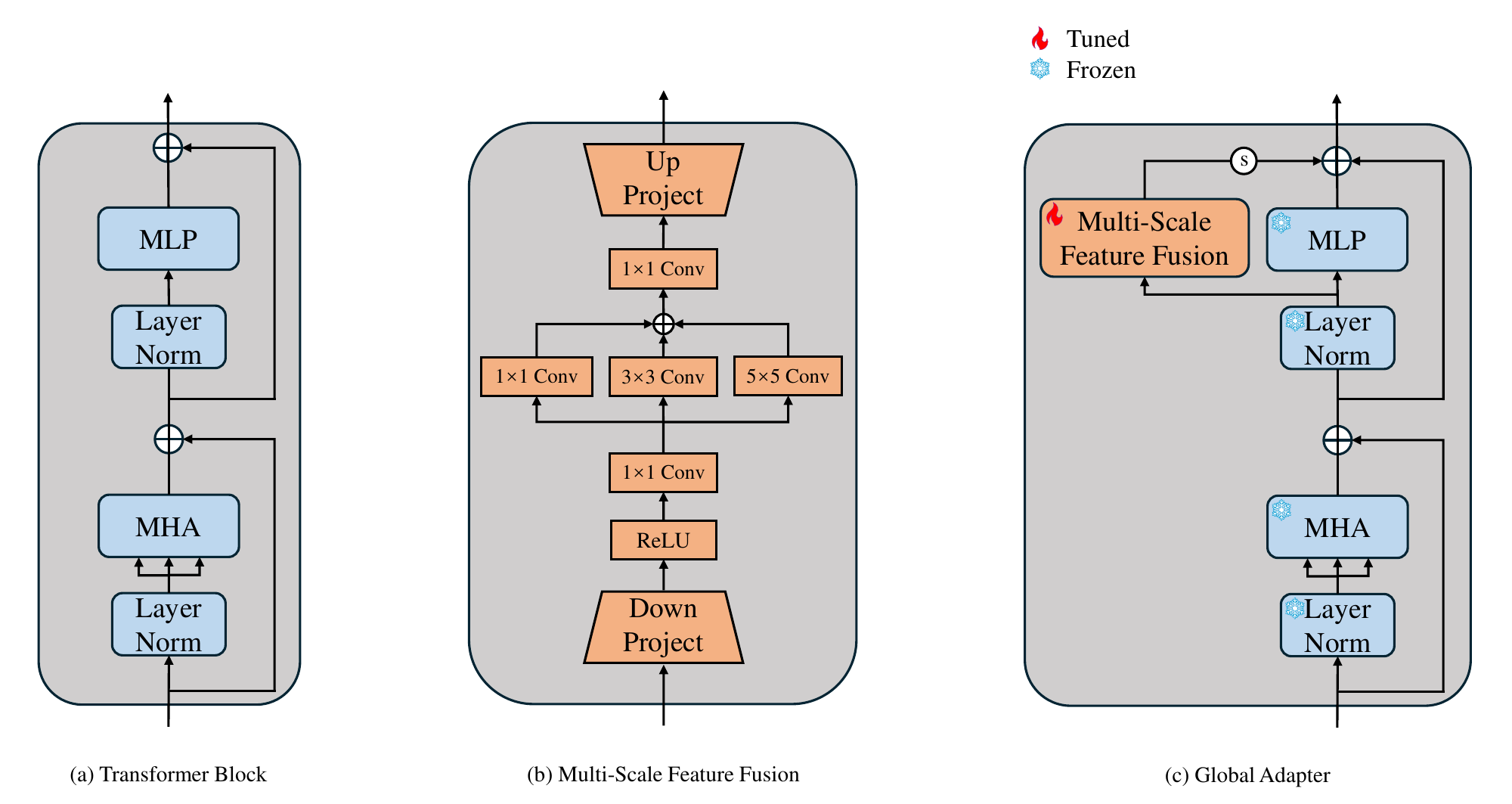}
\caption{\label{f2}Illustration of multi-scale feature fusion module. (a) is a standard Transformer block, and (b) is the structure of a multi-scale feature fusion module. We parallelize the multi-scale feature fusion module with the MLP layer in each standard Transformer block to obtain the global adapter (c).}
\end{figure*}

\section{Related Works}
\textbf{Visual Place Recognition:} Early VPR methods used handcrafted local features that can be further aggregated into a global descriptor to represent the entire image, such as Fisher Vectors \cite{jegou2011aggregating}, Bag of Words \cite{philbin2007object}, and Local Aggregation Descriptor Vector (VLAD) \cite{jegou2010aggregating}, and used such global descriptors for retrieval to find the closest position to the query image. With the significant progress made in deep learning, current VPR methods \cite{wang2022transvpr, arandjelovic2016netvlad, ali2023mixvpr, keetha2023anyloc} mainly use CNN or transformer as the backbone network. At the same time, a series of aggregation methods for image feature descriptors \cite{zhu2023r2former, ali2023mixvpr, radenovic2018fine} have emerged, which either use direct query sorting for retrieval or are divided into two steps, with the first step being to retrieve a part of similar images and then reranking these images. Our method uses a one-step query approach to retrieve localization images, and it is worth noting that even if our model does not include reranking stages, we outperform all baselines using the two-stage method (thus much faster). Recently, there have also been works that view VPR tasks as image classification tasks \cite{berton2022rethinking}, solving the problem of training time scalability through the use of contrastive learning methods, allowing for learning from large-scale databases and achieving state-of-the-art results on many datasets.

A recent work \cite{izquierdo2023optimal} used the optimal transport algorithm \cite{cuturi2013sinkhorn} to optimize the allocation of local descriptors in clusters of images, and then performed one-step retrieval by discarding the unstable features of the images. This method easily discards some robust information as useless information, which is detrimental to the performance of the model. Another work \cite{lu2024towards} added some lightweight adapters to the pre-trained model to the pre-trained backbone, and fine-tuned the model to make the pre-trained model perceive robust information as image representation, thereby improving the robustness of the VPR model.

\textbf{Additional token extensions in transformers:} Extending transformer sequences with special tokens has become popular in BERT \cite{devlin2018bert}. However, most methods either add new tokens or provide new information to the network, such as [SEP] tokens in BERT and tape tokens in AdaTape \cite{xue2023adaptive}, or collect information from these tokens and use their output values as the output of the model. Recently, \cite{darcet2023vision} proposed a simple method to improve the transformer model by using memory tokens added to token sequences, which do not contain information and their output values are not used for any purpose. They are just registers, and the model can learn to store and retrieve information during forward propagation. Our method is inspired by this, and our work applies registers to the aggregation part, combined with fine-tuning of the pre-trained model to adapt to the changes of the foundation model in the VPR task, further retaining robust features and removing useless features during the aggregation process.

\section{Method}

Vision Transformer (ViT) \cite{dosovitskiy2020image} and its variants \cite{oquab2023dinov2} have been proven to be very powerful for various computer vision tasks, including VPR. In our work, we use a pre-trained DINOv2 model based on ViT for VPR tasks, which is consistent with the model that \cite{darcet2023vision} focuses on.

\subsection{Local descriptor extraction}
Given an input image, the ViT model will initially divide input image $I \in \mathbb{R}^{h \times w \times c}$ into $p \times p$ patches, where $p = 14$. These patches are sequentially passed through the transformer to generate output tokens \{$t_1,...,t_n,t_{n+1}$\}, $t_i \in \mathbb{R}^d$. Here $n=hw/p^2$ is the number of input patches, and $t_{n+1}$ is an additional learnable global token, represented by $[class]$. Its purpose is to capture the semantic information of the entire patch sequence, which aids the model in better comprehending the semantic content of the entire input sequence.

Before being fed into the transformer block, $n+1$ output tokens are first adding positional embeddings to preserve the positional information, and then fed into the transformer block to generate feature representations of image patches. The standard transformer block mainly includes Multiple Head Attention (MHA), Feedforward Neural Network (FFN), and LayerNormalization (LN) layers. The processing of the input token sequence can be divided into two parts. In the first part, the sequence undergoes three distinct linear transformations (Query, Key, and Value), computes the similarity score between Query and Key, converts the score into weights using the softmax function, multiplies the weights with Value to obtain a self-attention representation, and finally performs weighted summation and layer normalization. In the second part, the normalized results from the first part are first processed through a feedforward neural network, then subjected to weighted summation and another layer normalization to yield the final output. Repeating this transformer block multiple times enhances the model's representation capabilities. The process for a single block can be described as follows:
\begin{equation}
X_{\rm n}'={MHA(LN(X_{\rm n-1}))+X_{\rm n-1}}
\end{equation}
\begin{equation}
X_{\rm n}={MLP(LN(X_{\rm n}'))+X_{\rm n}'}
\end{equation}
Here $X_{n-1}$ and $Xn$ are the outputs of the ($n-1$)-th and $n$-th layers of the transformer block, respectively.

Although pre-trained foundation models offer robust feature representations, their full potential is not realized in VPR due to the disparity between pre-training and the VPR task. To address this, drawing inspiration from the multi-scale convolution adapter \cite{lu2024cricavpr} and \cite{szegedy2015going}, we improved their methods and introduced a global adapter for multi-scale feature fusion to fine-tune the pre-trained model. We employed a multi-scale feature fusion approach to adjust the transformer block, as illustrated in Fig. \ref{f2} .

\begin{figure*}[!t]
\centering
\includegraphics[scale=.46]{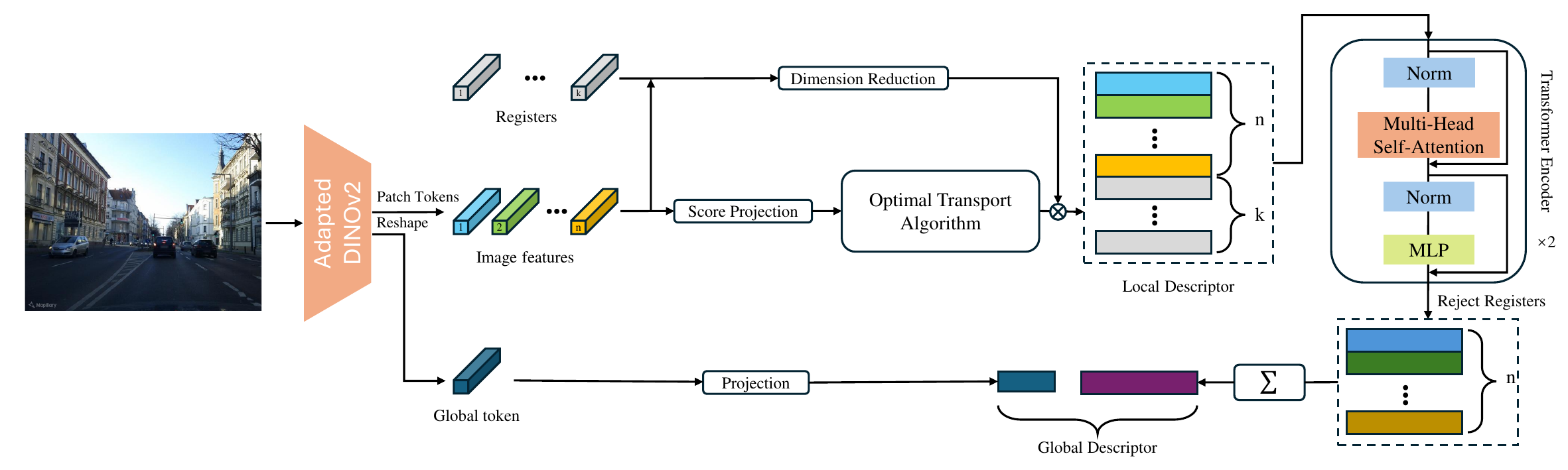}
\caption{\label{f3}Illustration of our VPR pipeline. Firstly, a ViT backbone with a multi-scale feature fusion module is used to extract local features and global labels, followed by score projection to obtain the score matrix for feature-to-cluster. Score projection is essentially a small MLP, and the optimal transport module uses the Sinkhorn algorithm. Then, we explicitly add registers to the sequence, which, along with local features, obtain local descriptors through a score matrix. At this point, the registers do not contain any information from the image. Then, the local descriptors with registers after dimensionality-reduction are fed into a Transformer Encoder with the aim of reallocating feature weights, assigning useless features to registers and discarding them. Finally, the remaining local descriptors are aggregated into the final descriptor and concatenated with the global token.}
\end{figure*}

Specifically, we incorporated a multi-scale feature fusion adapter within each transformer block, comprising an upsampling module, a downsampling module, and a channel-level fusion module. As the input token traverses the tail of a single transformer block, the output is downsampled and activated using ReLU. The downsampled output is then fed into the channel-level fusion module. The channel-level fusion module comprises three simple convolutions. The downsampled and activated features first undergo a $1\times1$ convolution to reduce the channel dimension, followed by convolutions of $1\times1$, $3\times3$, and $5\times5$ to extract features of varying scales. These features are then concatenated across channels to achieve feature fusion at the channel level, followed by a final $1\times1$ convolution to restore the output features to their original dimensionality. After passing through the channel-level feature fusion module, the features are multiplied by a scaling factor s and then upsampled. The resulting output is subsequently fed into the subsequent transformer block. This multi-scale feature fusion approach can be represented as follows:
\begin{equation}
X_{\rm n}'={MHA(LN(X_{\rm n-1}))+X_{\rm n-1}}
\end{equation}
\begin{equation}
X_{\rm n}={MLP(LN(X_{\rm n}'))+s\cdot MFF(LN(X_{\rm n}'))+X_{\rm n}'}
\end{equation}

We employ a global adapter to fine-tune the foundation model, enabling it to produce feature representations that are particularly attentive to static features while disregarding dynamic disturbances. This strategy effectively integrates the pre-trained foundation model into VPR tasks, significantly enhancing the model's performance.

\subsection{Register assisted aggregation}
Drawing on the findings in \cite{darcet2023vision}, during the aggregation of local descriptors, we classify features that are not intended to appear in the global descriptor as image artifacts. These artifacts align with those identified in \cite{darcet2023vision} and incorporate global background information. Excessive inclusion of these features in the global descriptor can degrade VPR performance. We introduce a novel aggregation method, as shown in the Fig. \ref{f3}: we retain the optimal transport \cite{cuturi2013sinkhorn} allocation of \cite{izquierdo2023optimal} features to clusters, along with score projection and dimensionality-reduction techniques. Unlike this approach, we eliminate the `Dustbin' from the optimal allocation matrix, as it contains static features that we prefer not to discard. Score projection can be formulated as:
\begin{equation}
s_{i}=W_{s_2}(\sigma(W_{s_1}(t_i)+b_{s_1}))+b_{s_2}
\end{equation}
where $W_{s_1}$ , $W_{s_2}$ and $b_{s_1}$ , $b_{s_2}$ are the weights and biases of the layers, and $\sigma$ is a non-linear activation function. Dimensionality-reduction can be expressed as:
\begin{equation}
f_{i}=W_{f_2}(\sigma(W_{f_1}(t_i)+b_{f_1}))+b_{f_2}
\end{equation}

In our approach, certain registers devoid of information are explicitly incorporated into the transformed feature sequence of the image input patch. Subsequently, these register-containing local feature descriptors undergo reassignment of feature weights and are subsequently discarded. The remaining local descriptors capture the desired robust retrieval information. Finally, these local descriptors are summed and concatenated with the global token to yield the image's final global descriptor. During concatenation, we employ the same scene descriptor $g$ as in \cite{izquierdo2023optimal}. This is because scene-related global information, which is not easily integrated into local features, is contained within g. The concatenation method is as follows:
\begin{equation}
g=W_{g_2}(\sigma(W_{g_1}(t_{n+1}))+b_{g_1}))+b_{g_2}
\end{equation}
where $t_{n+1}$ is the global token from DINOv2 after fine-tuning the global adapter.

To effectively store these artifacts in registers and discard them, we developed a module that simulates a transformer, termed the attention encoder. This module facilitates the removal of artifacts by incorporating local descriptors into the registers. Surprisingly, it effectively assigns features that contain extensive dynamic background information to the registers, which are precisely the features we wish to exclude from the global descriptors. We also conducted ablation experiments on the number of registers, The results of recall rate and the comparison of heatmaps demonstrate the effectiveness of our approach.

\section{Experiments}
\subsection{Implementation details}

\begin{table*}[!t]
\caption{\label{tab1}Comparison to state-of-the-art methods on benchmark datasets. The best is highlighted in bold and the second is underlined.}
\centering
\scalebox{0.83}{ 
\begin{tabular}{lccccccccccccccccccc}
\toprule[1pt]
\multicolumn{1}{c}{\multirow{2}{*}{Method}} & \multicolumn{3}{c}{MSLS Val}                  & \multicolumn{1}{l}{} & \multicolumn{3}{c}{Pitts250k-test}            &  & \multicolumn{3}{c}{Pitts30k-test}             &  & \multicolumn{3}{c}{NordLand}                  &                      & \multicolumn{3}{c}{SPED}                      \\ \cline{2-4} \cline{6-8} \cline{10-12} \cline{14-16} \cline{18-20} 
\multicolumn{1}{c}{}                        & R@1           & R@5           & R@10          &                      & R@1           & R@5           & R@10          &  & R@1           & R@5           & R@10          &  & R@1           & R@5           & R@10          & \multicolumn{1}{l}{} & R@1           & R@5           & R@10          \\ \hline
NetVLAD                                     & 82.6          & 89.6          & 92.0          &                      & 90.5          & 96.2          & 97.4          &  & 81.9          & 91.2          & 93.7          &  & 32.6          & 47.1          & 53.3          &                      & 78.7          & 88.3          & 91.4          \\
GeM                                         & 76.5          & 85.7          & 88.2          &                      & 82.9          & 92.1          & 94.3          &  & 80.5          & 91.8          & 96.2          &  & 20.8          & 33.3          & 40.0          &                      & 64.6          & 79.4          & 83.5          \\
CosPlace                                    & 84.5          & 90.1          & 91.8          &                      & 91.5          & 96.9          & 97.9          &  & 90.9          & 95.7          & 96.7          &  & 58.5          & 73.7          & 79.4          &                      & 75.3          & 85.9          & 88.6          \\
EigenPlaces                                 & 89.3          & 93.7          & 95.0          &                      & 94.1          & 98.0          & 98.7          &  & 92.5          & {\underline{96.8}}    & {\underline{97.6}}    &  & 54.4          & 68.8          & 74.1          &                      & 69.9          & 82.9          & 87.6          \\
MixVPR                                      & 88.0          & 92.7          & 94.6          &                      & \underline{94.6} & 98.3    & 99.0    &  & {\underline{91.5}}    & 95.5          & 96.3          &  & 58.4          & 74.6          & 80.0          &                      & 85.2          & 92.1          & 94.6          \\
SALAD                                       & {\underline{90.7}}    & {\underline{95.5}}    & {\underline{96.1}}    &                      & 94.2          & {\underline{98.4}}          & {\underline{99.1}}          &  & 91.1          & 96.3          & 97.2          &  & {\underline{74.4}}    & {\underline{88.2}}    & {\underline{91.3}}    &                      & {\underline{89.8}}    & {\underline{94.7}}    & {\underline{95.8}}    \\ \hline
\textbf{Ours}                               & \textbf{91.4} & \textbf{96.2} & \textbf{96.9} &                      & {\textbf{94.8}}    & \textbf{98.7} & \textbf{99.3} &  & \textbf{91.6} & \textbf{96.9} & \textbf{97.7} &  & \textbf{75.1} & \textbf{90.3} & \textbf{93.5} &                      & \textbf{90.2} & \textbf{95.1} & \textbf{96.5} \\ 
\bottomrule[1pt]
\end{tabular}
}
\end{table*}

\textbf{Dataset:} We trained all models on the same dataset according to the standard framework of GSV-Cities \cite{ali2022gsv}, which proposed a high-precision dataset of 67k locations depicted by 560k images. We evaluated the model on a benchmark of 5 datasets. Pitts250k-test \cite{torii2013visual}, which includes 8k queries and 83k reference images. Pitts30k-test \cite{torii2013visual} is a subset of Pitts250k, consisting of 8k queries and 8k references. Both Pittsburgh datasets show significant viewpoint changes. Mapillary Street Level Sequences (MSLS) \cite{warburg2020mapillary} consists of over 1.6 million images collected in urban, suburban, and natural scenes over the past 7 years. The SPED \cite{zaffar2021vpr} benchmark includes 607 queries and 607 references from surveillance cameras, showing significant seasonal and lighting changes. Nordland \cite{zaffar2021vpr} is a highly challenging benchmark that has been collected using cameras installed in front of trains for four seasons, covering scenes from snowy winter to sunny summer, with extreme changes in appearance.

\begin{figure*}[!t]
\centering
\includegraphics[scale=.5]{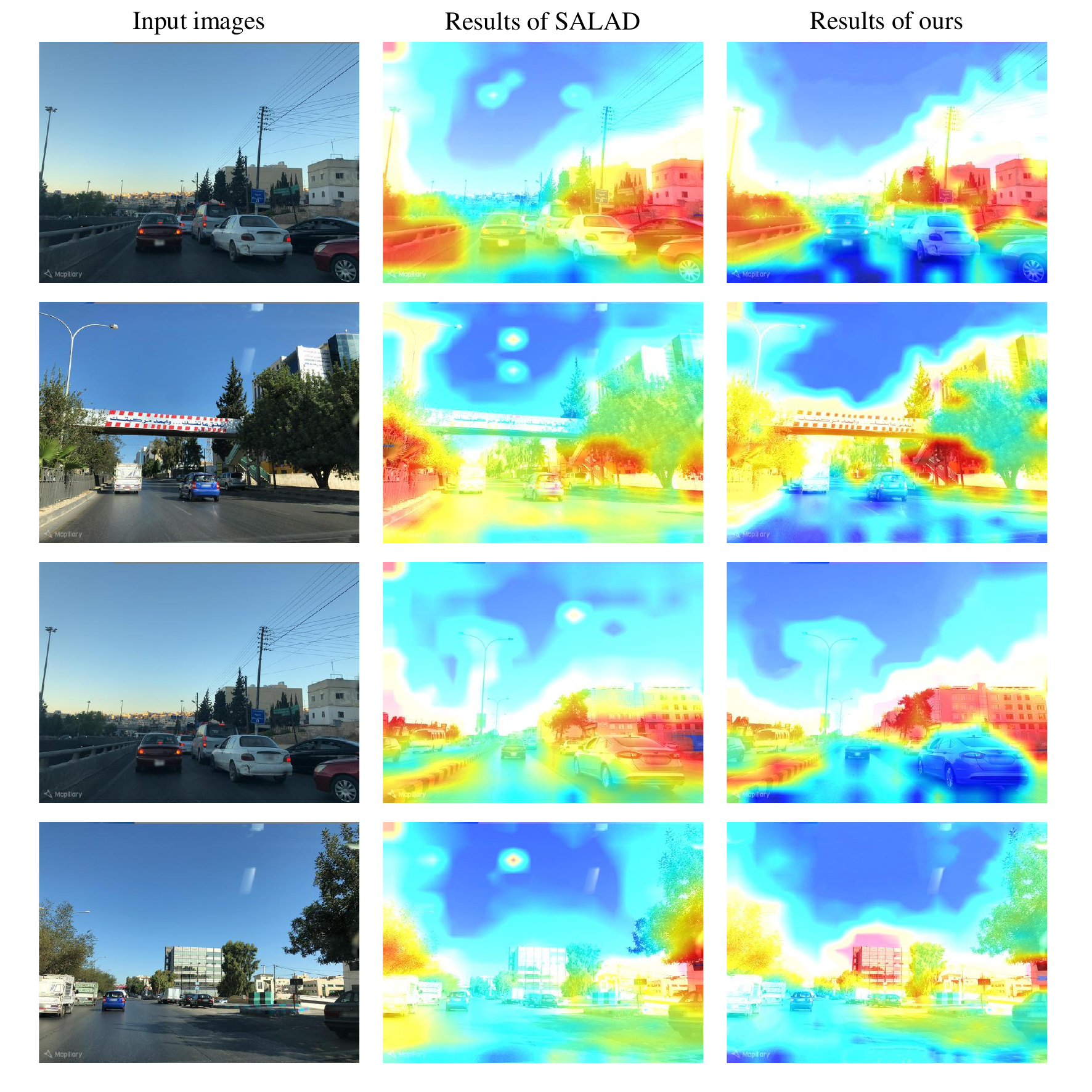}
\caption{\label{f4}Attention map visualizations of SALAD model and our model. We compute the mean in the channel dimension of the output feature map and display it using the heatmap. The feature map of the SALAD model may contain some features that are not helpful for VPR tasks, such as cars, and discard features that are helpful for retrieval, such as buildings and overpasses. Compared to the visual feature maps of SALAD in the sky and on the road, our method is smoother.}
\end{figure*}

\begin{figure*}[!t]
\centering
\includegraphics[scale=.4]{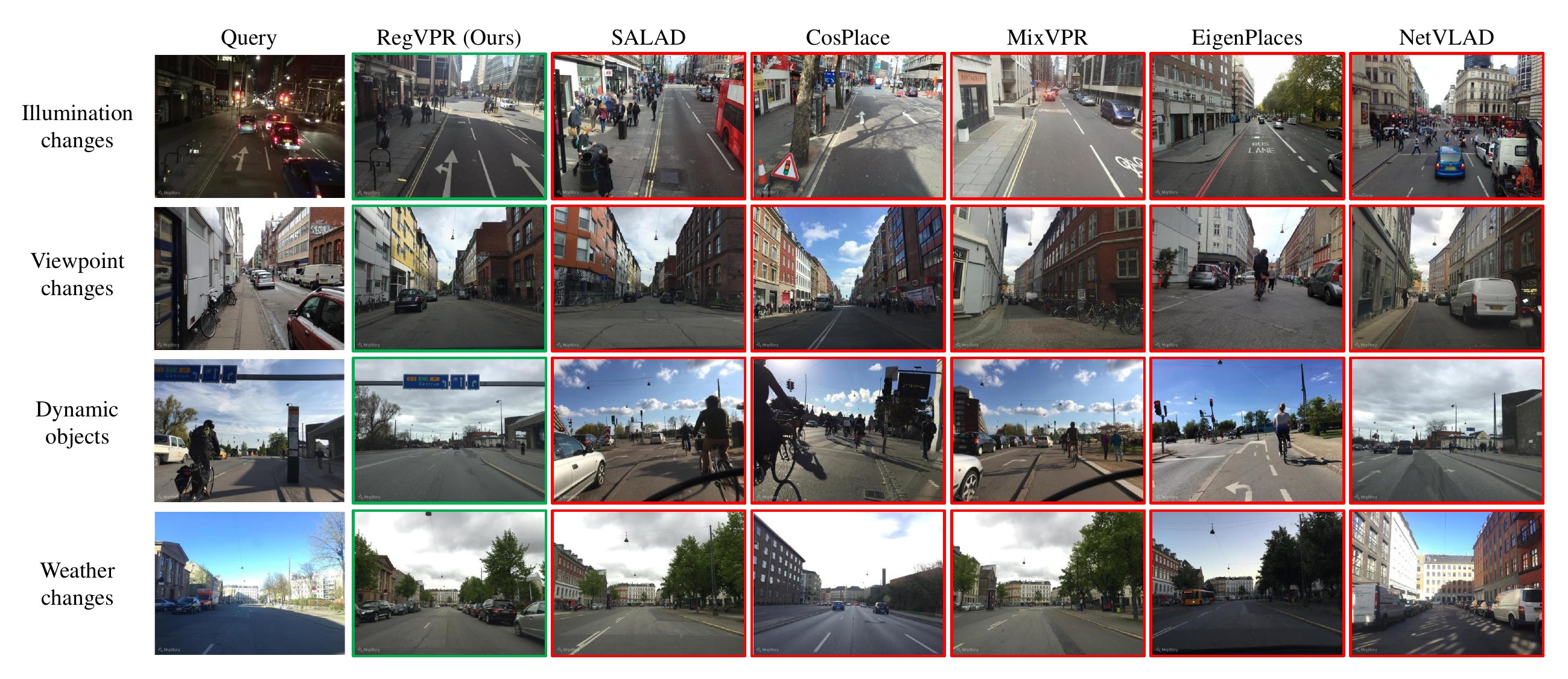}
\caption{\label{f5}Qualitative results. In these four challenging examples (including light changes, viewpoint changes, dynamic objects, and weather changes), our method successfully retrieved the correct database images, while all other methods produced incorrect results.}
\end{figure*}

\textbf{Structure:} Our method, implemented in the PyTorch framework \cite{paszke2019pytorch}, uses a pre-trained DINOv2 backbone \cite{oquab2023dinov2} on ImageNet \cite{krizhevsky2012imagenet}, and the chosen version is Vit-B/14. The input image resolution is $224\times224$, and the backbone token dimension is 768. The scaling factor $s$ in the formula is set to 0.2, and the bottleneck ratio of the multi-scale feature fusion module is set to 0.5, so the convolution input dimension in channel level feature fusion is 384. For the fully connected layer, the weights of hidden layers $W_{s_1}$, $W_{f_1}$, and $W_{g_1}$ have 512 neurons, and ReLU is used as the activation function. In order to improve computational efficiency, we adopt dimensionality reduction by compressing the feature and global label dimensions from 768 to 128. For the optimal transport algorithm \cite{cuturi2013sinkhorn}, we use $m=64$ clusters, and the final global descriptor size is $128\times64+256=8448$.

\textbf{Training:} We trained for 2 hours on a single NVIDIA 3080Ti. For the loss function, we use multiple similarity loss \cite{wang2019multi} as it has been proven to perform best in VPR tasks. We use batches with $P=60$ places, each batch described by 4 images. We optimized using AdamW \cite{loshchilov2017decoupled} with an initial learning rate of 6e-5. We use a dropout of 0.3 on fractional projection and dimensionality reduction neurons. Finally, we use images scaled to $224\times224$ for training up to 4 epochs. In model training, we define potential positive images as reference images within 10 meters of the query image, while determined negative images are those that exceed 25 meters of reference images. We follow the same evaluation criteria, where measurement $Recall@k$ ($R@k$) . If at least one of the first $k$ reference images retrieved is within 25 meters of the query image, it is determined that the query image has been successfully retrieved.

\subsection{Quantitative results}
Table \ref{tab1} shows the quantitative results of our method compared to several single-stage methods, including two traditional baselines, NetVLAD \cite{arandjelovic2016netvlad} and GeM \cite{radenovic2018fine}, as well as the most recent best performing baselines, CosPlace \cite{berton2022rethinking}, EigenPlaces \cite{berton2023eigenplaces}, MixVPR \cite{ali2023mixvpr}, and SALAD \cite{izquierdo2023optimal}. The dataset we used in the evaluation phase includes MSLS Validation, Pitts250k-test, Pitts30k-test, NordLand, and SPED. Please note that the evaluation results of SALAD \cite{izquierdo2023optimal} were reproduced locally using the code provided by the author. Our method achieved the best $R@1$, $R@5$, and $R@10$ results on the dataset used for evaluation.

Compared with SALAD \cite{izquierdo2023optimal}, our results are outstanding, especially in the highly challenging evaluation of NordLand, with improvements of 2.1\% and 2.2\% in $R@5$ and $R@10$, respectively. The main reason for this improvement is that our method can generate more comprehensive coverage of robust features in the image, and also effectively filter out useless global background information.

\subsection{Qualitative results}
We compared the SALAD \cite{izquierdo2023optimal} model with our model in terms of feature weight allocation by creating a heatmap, with the results presented in Fig. \ref{f4}. The figure clearly demonstrates that the SALAD method discards certain features with robust representations. In contrast, our method successfully retains these features and does not overly focus on global background features or dynamic non-robust features.

Additionally, we conducted retrieval experiments with several other methods in extreme environments, considering challenges such as lighting, viewpoints, dynamic objects, and weather changes. The results are presented in Fig. \ref{f5}. Our method accurately retrieves the image most closely related to the query image, whereas other methods either retrieve highly similar images but with a significant positional distance, or retrieve images whose positional distance exceeds our set threshold. This demonstrates the robustness of our approach.

\subsection{Ablation study}
In this section, we conducted a series of ablation experiments to verify the necessity of fine-tuning the backbone network and the effectiveness of our proposed aggregation method.
\subsubsection{Fine-tuning the network}
Based on fine-tuning the DINOv2 \cite{oquab2023dinov2} backbone network using a global adapter, we compared our local feature aggregation method with other aggregation methods. As presented in the Table \ref{tab2}, we observed that the pre-trained DINOv2 backbone network fine-tuned with a global adapter outperformed models that freeze the first 8 layers and only train the last 4 layers in all aggregation methods. Additionally, our aggregation method surpassed models trained in the same manner for the SALAD \cite{izquierdo2023optimal} when only training the last four layers of the DINOv2 backbone network, demonstrating the efficacy of our register aggregation approach. We also observed a curious phenomenon: models trained with the SALAD aggregation method and a global adapter to fine-tune the last four layers showed a negative improvement in R@1 results across both datasets. We hypothesize that this phenomenon is due to the fact that fine-tuning the backbone network with a global adapter enhances feature extraction, whereas the SALAD's Dustbin method discards more robust features that are beneficial for retrieval when only training the last four layers.

\begin{table}[!t]
\caption{\label{tab2}Ablation experiments. The best is highlighted in bold and the second is underlined.}
\centering
\scalebox{0.65}{
\begin{tabular}{clcccccccc}
\toprule[1pt]
\multicolumn{2}{c}{\multirow{2}{*}{Ablated versions}}                                                    &  & \multicolumn{3}{c}{MSLS Val}                  &  & \multicolumn{3}{c}{NordLand}                  \\ \cline{4-6} \cline{8-10} 
\multicolumn{2}{c}{}                                                                                     &  & R@1           & R@5           & R@10          &  & R@1           & R@5           & R@10          \\ \hline
\multirow{3}{*}{\begin{tabular}[c]{@{}c@{}}DINOv2\\ (Frozen)\end{tabular}}              & +GeM           &  & 44.6          & 55.9          & 59.2          &  & 17.7          & 30.6          & 38.5          \\
                                                                                        & +SALAD         &  & 88.0          & 93.9          & 95.0          &  & 70.4          & 83.5          & 88.2          \\
                                                                                        & \textbf{+Ours} &  & 88.3          & 95.3          & 96.2          &  & 71.1          & 85.7          & 90.1          \\ \hline
\multirow{3}{*}{\begin{tabular}[c]{@{}c@{}}DINOv2\\ (Train last 4 blocks)\end{tabular}} & +GeM           &  & 83.9          & 90.3          & 94.5          &  & 35.1          & 51.9          & 58.8          \\
                                                                                        & +SALAD         &  & 90.7          & 95.5          & 96.1          &  & 74.4          & 88.2          & 91.3          \\
                                                                                        & \textbf{+Ours} &  & {\underline{90.8}}    & {\underline{96.1}}    & {\underline{96.8}}    &  & {\underline{74.5}}    & {\underline{89.8}}    & {\underline{93.2}}    \\ \hline
\multirow{3}{*}{\begin{tabular}[c]{@{}c@{}}DINOv2\\ (Global Adapter)\end{tabular}}      & +GeM           &  & 82.8          & 91.6          & 93.2          &  & 37.3          & 55.2          & 62.8          \\
                                                                                        & +SALAD         &  & 90.5          & 95.7          & 96.2          &  & 74.3          & 88.9          & 92.0          \\
                                                                                        & \textbf{+Ours} &  & \textbf{91.4} & \textbf{96.2} & \textbf{96.9} &  & \textbf{75.1} & \textbf{90.3} & \textbf{93.5} \\ \bottomrule[1pt]
\end{tabular}
}
\end{table}

\subsubsection{Hyperparameter}
\begin{figure}[!t]
\centering
\includegraphics[scale=.34]{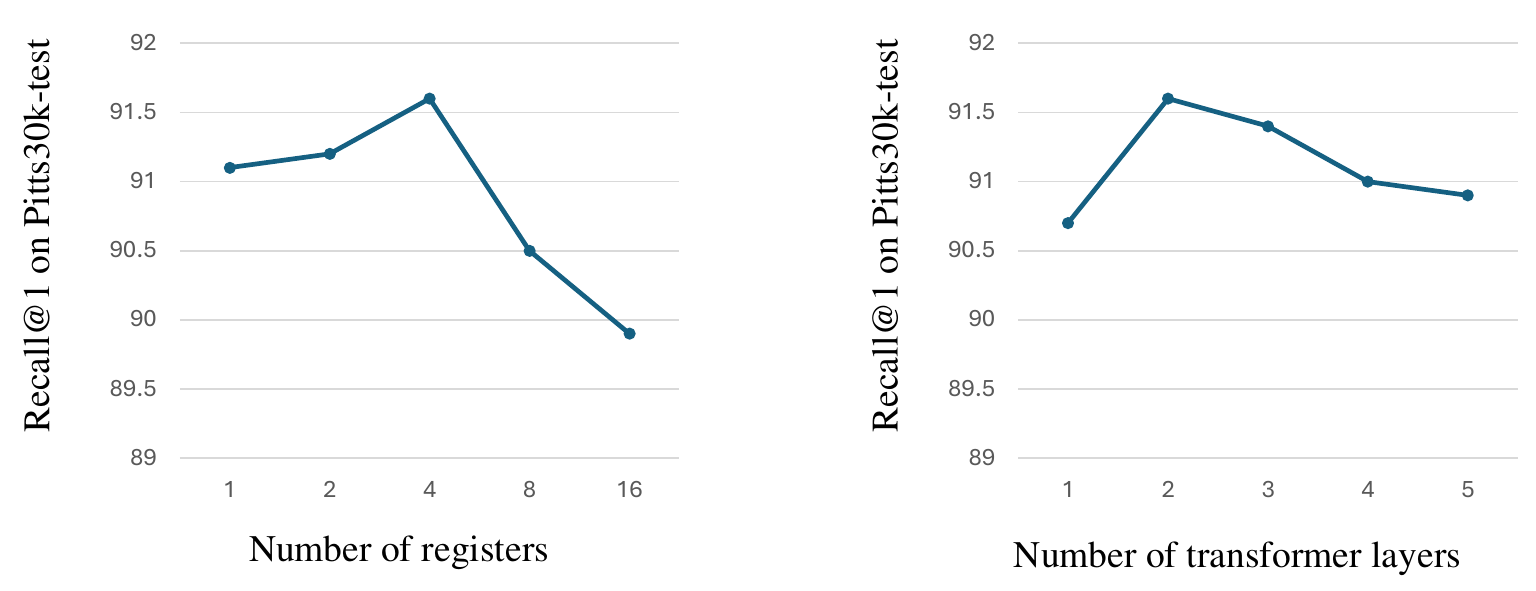}
\caption{\label{f6}Ablation of the the number of registers and Transformer Encoder layers. (left): When the number of registers is set to 4, the model reaches its highest performance, and having more registers does not lead to better retrieval performance. (right): When the number of layers in the Transformer Encoder is set to 2, the model achieves optimal reassignment of feature weights.}
\end{figure}

To demonstrate the impact of registers on model performance, we evaluated our model on the MSLS Validation and Pitts30k-test datasets, varying the number of registers and the layers of the Transformer Encoder. Based on fine-tuning the DINOv2 backbone network, we trained models with 1, 2, 4, 8, or 16 registers. The left of Fig. \ref{f6} illustrates the impact of different register counts on model performance. Observing the quantitative results, it is evident that the model performs optimally with 4 registers. We also conducted an ablation study on the number of layers in the Transformer Encoder, as illustrated in the right of Fig. \ref{f6}. While maintaining the optimal number of registers, the model performs best with a Transformer Encoder consisting of 2 layers.

\section{Conclusion and limitations}
In this study, we introduced a novel register assisted aggregation technique that combines local features extracted from pre-trained networks with registers. Following a simulation of the Transformer Encoder, non-robust features, which are rich in global background information, are filtered out, resulting in a robust global descriptor. During the feature extraction phase, we also fine-tuned the pre-trained network for the VPR task. Our experimental findings demonstrated that our aggregation approach surpasses existing benchmarks, outperforming even some two-stage retrieval techniques. Extensive ablation experiments have confirmed the effectiveness of each module.

Limitations: In terms of experimental results, we observed minimal improvements in $R@1$ relative to $R@5$ and $R@10$. Our analysis suggests that while the model focuses on robust features beyond the global context, these distinctions are not significant for features with spatial information, leading to perceptual aliasing. This is a common issue in first-stage VPR methods, which we will further investigate in our future work.

\bibliographystyle{model1-num-names.bst}
\bibliography{refs}

\begin{thebibliography}{36}
\expandafter\ifx\csname natexlab\endcsname\relax\def\natexlab#1{#1}\fi
\providecommand{\url}[1]{\texttt{#1}}
\providecommand{\href}[2]{#2}
\providecommand{\path}[1]{#1}
\providecommand{\DOIprefix}{doi:}
\providecommand{\ArXivprefix}{arXiv:}
\providecommand{\URLprefix}{URL: }
\providecommand{\Pubmedprefix}{pmid:}
\providecommand{\doi}[1]{\href{http://dx.doi.org/#1}{\path{#1}}}
\providecommand{\Pubmed}[1]{\href{pmid:#1}{\path{#1}}}
\providecommand{\bibinfo}[2]{#2}
\ifx\xfnm\relax \def\xfnm[#1]{\unskip,\space#1}\fi
\bibitem[{Xu et~al.(2020)Xu, Snderhauf, and Milford}]{xu2020probabilistic}
\bibinfo{author}{M.~Xu}, \bibinfo{author}{N.~Snderhauf}, \bibinfo{author}{M.~Milford},
\newblock \bibinfo{title}{Probabilistic visual place recognition for hierarchical localization},
\newblock \bibinfo{journal}{IEEE Robotics and Automation Letters} \bibinfo{volume}{6} (\bibinfo{year}{2020}) \bibinfo{pages}{311--318}.
\bibitem[{Middelberg et~al.(2014)Middelberg, Sattler, Untzelmann, and Kobbelt}]{middelberg2014scalable}
\bibinfo{author}{S.~Middelberg}, \bibinfo{author}{T.~Sattler}, \bibinfo{author}{O.~Untzelmann}, \bibinfo{author}{L.~Kobbelt},
\newblock \bibinfo{title}{Scalable 6-dof localization on mobile devices},
\newblock in: \bibinfo{booktitle}{Computer Vision--ECCV 2014: 13th European Conference, Zurich, Switzerland, September 6-12, 2014, Proceedings, Part II 13}, \bibinfo{organization}{Springer}, \bibinfo{year}{2014}, pp. \bibinfo{pages}{268--283}.
\bibitem[{Doan et~al.(2019)Doan, Latif, Chin, Liu, Do, and Reid}]{doan2019scalable}
\bibinfo{author}{A.-D. Doan}, \bibinfo{author}{Y.~Latif}, \bibinfo{author}{T.-J. Chin}, \bibinfo{author}{Y.~Liu}, \bibinfo{author}{T.-T. Do}, \bibinfo{author}{I.~Reid},
\newblock \bibinfo{title}{Scalable place recognition under appearance change for autonomous driving},
\newblock in: \bibinfo{booktitle}{Proceedings of the IEEE/CVF International Conference on Computer Vision}, \bibinfo{year}{2019}, pp. \bibinfo{pages}{9319--9328}.
\bibitem[{Liu et~al.(2019{\natexlab{a}})Liu, Li, and Dai}]{liu2019stochastic}
\bibinfo{author}{L.~Liu}, \bibinfo{author}{H.~Li}, \bibinfo{author}{Y.~Dai},
\newblock \bibinfo{title}{Stochastic attraction-repulsion embedding for large scale image localization},
\newblock in: \bibinfo{booktitle}{Proceedings of the IEEE/CVF International Conference on Computer Vision}, \bibinfo{year}{2019}{\natexlab{a}}, pp. \bibinfo{pages}{2570--2579}.
\bibitem[{Liu et~al.(2019{\natexlab{b}})Liu, Zhou, Suo, Yin, Chen, Wang, Li, and Liu}]{liu2019lpd}
\bibinfo{author}{Z.~Liu}, \bibinfo{author}{S.~Zhou}, \bibinfo{author}{C.~Suo}, \bibinfo{author}{P.~Yin}, \bibinfo{author}{W.~Chen}, \bibinfo{author}{H.~Wang}, \bibinfo{author}{H.~Li}, \bibinfo{author}{Y.-H. Liu},
\newblock \bibinfo{title}{Lpd-net: 3d point cloud learning for large-scale place recognition and environment analysis},
\newblock in: \bibinfo{booktitle}{Proceedings of the IEEE/CVF international conference on computer vision}, \bibinfo{year}{2019}{\natexlab{b}}, pp. \bibinfo{pages}{2831--2840}.
\bibitem[{J{\'e}gou et~al.(2010)J{\'e}gou, Douze, Schmid, and P{\'e}rez}]{jegou2010aggregating}
\bibinfo{author}{H.~J{\'e}gou}, \bibinfo{author}{M.~Douze}, \bibinfo{author}{C.~Schmid}, \bibinfo{author}{P.~P{\'e}rez},
\newblock \bibinfo{title}{Aggregating local descriptors into a compact image representation},
\newblock in: \bibinfo{booktitle}{2010 IEEE computer society conference on computer vision and pattern recognition}, \bibinfo{organization}{IEEE}, \bibinfo{year}{2010}, pp. \bibinfo{pages}{3304--3311}.
\bibitem[{J{\'e}gou et~al.(2011)J{\'e}gou, Perronnin, Douze, S{\'a}nchez, P{\'e}rez, and Schmid}]{jegou2011aggregating}
\bibinfo{author}{H.~J{\'e}gou}, \bibinfo{author}{F.~Perronnin}, \bibinfo{author}{M.~Douze}, \bibinfo{author}{J.~S{\'a}nchez}, \bibinfo{author}{P.~P{\'e}rez}, \bibinfo{author}{C.~Schmid},
\newblock \bibinfo{title}{Aggregating local image descriptors into compact codes},
\newblock \bibinfo{journal}{IEEE transactions on pattern analysis and machine intelligence} \bibinfo{volume}{34} (\bibinfo{year}{2011}) \bibinfo{pages}{1704--1716}.
\bibitem[{Kim et~al.(2015)Kim, Dunn, and Frahm}]{kim2015predicting}
\bibinfo{author}{H.~J. Kim}, \bibinfo{author}{E.~Dunn}, \bibinfo{author}{J.-M. Frahm},
\newblock \bibinfo{title}{Predicting good features for image geo-localization using per-bundle vlad},
\newblock in: \bibinfo{booktitle}{Proceedings of the IEEE International Conference on Computer Vision}, \bibinfo{year}{2015}, pp. \bibinfo{pages}{1170--1178}.
\bibitem[{Vaswani et~al.(2017)Vaswani, Shazeer, Parmar, Uszkoreit, Jones, Gomez, Kaiser, and Polosukhin}]{vaswani2017attention}
\bibinfo{author}{A.~Vaswani}, \bibinfo{author}{N.~Shazeer}, \bibinfo{author}{N.~Parmar}, \bibinfo{author}{J.~Uszkoreit}, \bibinfo{author}{L.~Jones}, \bibinfo{author}{A.~N. Gomez}, \bibinfo{author}{{\L}.~Kaiser}, \bibinfo{author}{I.~Polosukhin},
\newblock \bibinfo{title}{Attention is all you need},
\newblock \bibinfo{journal}{Advances in neural information processing systems} \bibinfo{volume}{30} (\bibinfo{year}{2017}).
\bibitem[{Wang et~al.(2022)Wang, Shen, Zuo, Zhou, and Zheng}]{wang2022transvpr}
\bibinfo{author}{R.~Wang}, \bibinfo{author}{Y.~Shen}, \bibinfo{author}{W.~Zuo}, \bibinfo{author}{S.~Zhou}, \bibinfo{author}{N.~Zheng},
\newblock \bibinfo{title}{Transvpr: Transformer-based place recognition with multi-level attention aggregation},
\newblock in: \bibinfo{booktitle}{Proceedings of the IEEE/CVF Conference on Computer Vision and Pattern Recognition}, \bibinfo{year}{2022}, pp. \bibinfo{pages}{13648--13657}.
\bibitem[{Zhu et~al.(2023)Zhu, Yang, Chen, Shah, Shen, and Wang}]{zhu2023r2former}
\bibinfo{author}{S.~Zhu}, \bibinfo{author}{L.~Yang}, \bibinfo{author}{C.~Chen}, \bibinfo{author}{M.~Shah}, \bibinfo{author}{X.~Shen}, \bibinfo{author}{H.~Wang},
\newblock \bibinfo{title}{R2former: Unified retrieval and reranking transformer for place recognition},
\newblock in: \bibinfo{booktitle}{Proceedings of the IEEE/CVF Conference on Computer Vision and Pattern Recognition}, \bibinfo{year}{2023}, pp. \bibinfo{pages}{19370--19380}.
\bibitem[{Izquierdo and Civera(2023)}]{izquierdo2023optimal}
\bibinfo{author}{S.~Izquierdo}, \bibinfo{author}{J.~Civera},
\newblock \bibinfo{title}{Optimal transport aggregation for visual place recognition},
\newblock \bibinfo{journal}{arXiv preprint arXiv:2311.15937}  (\bibinfo{year}{2023}).
\bibitem[{Lu et~al.(2024)Lu, Zhang, Lan, Dong, Wang, and Yuan}]{lu2024towards}
\bibinfo{author}{F.~Lu}, \bibinfo{author}{L.~Zhang}, \bibinfo{author}{X.~Lan}, \bibinfo{author}{S.~Dong}, \bibinfo{author}{Y.~Wang}, \bibinfo{author}{C.~Yuan},
\newblock \bibinfo{title}{Towards seamless adaptation of pre-trained models for visual place recognition},
\newblock \bibinfo{journal}{arXiv preprint arXiv:2402.14505}  (\bibinfo{year}{2024}).
\bibitem[{Arandjelovic et~al.(2016)Arandjelovic, Gronat, Torii, Pajdla, and Sivic}]{arandjelovic2016netvlad}
\bibinfo{author}{R.~Arandjelovic}, \bibinfo{author}{P.~Gronat}, \bibinfo{author}{A.~Torii}, \bibinfo{author}{T.~Pajdla}, \bibinfo{author}{J.~Sivic},
\newblock \bibinfo{title}{Netvlad: Cnn architecture for weakly supervised place recognition},
\newblock in: \bibinfo{booktitle}{Proceedings of the IEEE conference on computer vision and pattern recognition}, \bibinfo{year}{2016}, pp. \bibinfo{pages}{5297--5307}.
\bibitem[{Darcet et~al.(2023)Darcet, Oquab, Mairal, and Bojanowski}]{darcet2023vision}
\bibinfo{author}{T.~Darcet}, \bibinfo{author}{M.~Oquab}, \bibinfo{author}{J.~Mairal}, \bibinfo{author}{P.~Bojanowski},
\newblock \bibinfo{title}{Vision transformers need registers},
\newblock \bibinfo{journal}{arXiv preprint arXiv:2309.16588}  (\bibinfo{year}{2023}).
\bibitem[{Oquab et~al.(2023)Oquab, Darcet, Moutakanni, Vo, Szafraniec, Khalidov, Fernandez, Haziza, Massa, El-Nouby et~al.}]{oquab2023dinov2}
\bibinfo{author}{M.~Oquab}, \bibinfo{author}{T.~Darcet}, \bibinfo{author}{T.~Moutakanni}, \bibinfo{author}{H.~Vo}, \bibinfo{author}{M.~Szafraniec}, \bibinfo{author}{V.~Khalidov}, \bibinfo{author}{P.~Fernandez}, \bibinfo{author}{D.~Haziza}, \bibinfo{author}{F.~Massa}, \bibinfo{author}{A.~El-Nouby}, et~al.,
\newblock \bibinfo{title}{Dinov2: Learning robust visual features without supervision},
\newblock \bibinfo{journal}{arXiv preprint arXiv:2304.07193}  (\bibinfo{year}{2023}).
\bibitem[{Philbin et~al.(2007)Philbin, Chum, Isard, Sivic, and Zisserman}]{philbin2007object}
\bibinfo{author}{J.~Philbin}, \bibinfo{author}{O.~Chum}, \bibinfo{author}{M.~Isard}, \bibinfo{author}{J.~Sivic}, \bibinfo{author}{A.~Zisserman},
\newblock \bibinfo{title}{Object retrieval with large vocabularies and fast spatial matching},
\newblock in: \bibinfo{booktitle}{2007 IEEE conference on computer vision and pattern recognition}, \bibinfo{organization}{IEEE}, \bibinfo{year}{2007}, pp. \bibinfo{pages}{1--8}.
\bibitem[{Ali-Bey et~al.(2023)Ali-Bey, Chaib-Draa, and Giguere}]{ali2023mixvpr}
\bibinfo{author}{A.~Ali-Bey}, \bibinfo{author}{B.~Chaib-Draa}, \bibinfo{author}{P.~Giguere},
\newblock \bibinfo{title}{Mixvpr: Feature mixing for visual place recognition},
\newblock in: \bibinfo{booktitle}{Proceedings of the IEEE/CVF Winter Conference on Applications of Computer Vision}, \bibinfo{year}{2023}, pp. \bibinfo{pages}{2998--3007}.
\bibitem[{Keetha et~al.(2023)Keetha, Mishra, Karhade, Jatavallabhula, Scherer, Krishna, and Garg}]{keetha2023anyloc}
\bibinfo{author}{N.~Keetha}, \bibinfo{author}{A.~Mishra}, \bibinfo{author}{J.~Karhade}, \bibinfo{author}{K.~M. Jatavallabhula}, \bibinfo{author}{S.~Scherer}, \bibinfo{author}{M.~Krishna}, \bibinfo{author}{S.~Garg},
\newblock \bibinfo{title}{Anyloc: Towards universal visual place recognition},
\newblock \bibinfo{journal}{IEEE Robotics and Automation Letters}  (\bibinfo{year}{2023}).
\bibitem[{Radenovi{\'c} et~al.(2018)Radenovi{\'c}, Tolias, and Chum}]{radenovic2018fine}
\bibinfo{author}{F.~Radenovi{\'c}}, \bibinfo{author}{G.~Tolias}, \bibinfo{author}{O.~Chum},
\newblock \bibinfo{title}{Fine-tuning cnn image retrieval with no human annotation},
\newblock \bibinfo{journal}{IEEE transactions on pattern analysis and machine intelligence} \bibinfo{volume}{41} (\bibinfo{year}{2018}) \bibinfo{pages}{1655--1668}.
\bibitem[{Berton et~al.(2022)Berton, Masone, and Caputo}]{berton2022rethinking}
\bibinfo{author}{G.~Berton}, \bibinfo{author}{C.~Masone}, \bibinfo{author}{B.~Caputo},
\newblock \bibinfo{title}{Rethinking visual geo-localization for large-scale applications},
\newblock in: \bibinfo{booktitle}{Proceedings of the IEEE/CVF Conference on Computer Vision and Pattern Recognition}, \bibinfo{year}{2022}, pp. \bibinfo{pages}{4878--4888}.
\bibitem[{Cuturi(2013)}]{cuturi2013sinkhorn}
\bibinfo{author}{M.~Cuturi},
\newblock \bibinfo{title}{Sinkhorn distances: Lightspeed computation of optimal transport},
\newblock \bibinfo{journal}{Advances in neural information processing systems} \bibinfo{volume}{26} (\bibinfo{year}{2013}).
\bibitem[{Devlin et~al.(2018)Devlin, Chang, Lee, and Toutanova}]{devlin2018bert}
\bibinfo{author}{J.~Devlin}, \bibinfo{author}{M.-W. Chang}, \bibinfo{author}{K.~Lee}, \bibinfo{author}{K.~Toutanova},
\newblock \bibinfo{title}{Bert: Pre-training of deep bidirectional transformers for language understanding},
\newblock \bibinfo{journal}{arXiv preprint arXiv:1810.04805}  (\bibinfo{year}{2018}).
\bibitem[{Xue et~al.(2023)Xue, Likhosherstov, Arnab, Houlsby, Dehghani, and You}]{xue2023adaptive}
\bibinfo{author}{F.~Xue}, \bibinfo{author}{V.~Likhosherstov}, \bibinfo{author}{A.~Arnab}, \bibinfo{author}{N.~Houlsby}, \bibinfo{author}{M.~Dehghani}, \bibinfo{author}{Y.~You},
\newblock \bibinfo{title}{Adaptive computation with elastic input sequence},
\newblock in: \bibinfo{booktitle}{International Conference on Machine Learning}, \bibinfo{organization}{PMLR}, \bibinfo{year}{2023}, pp. \bibinfo{pages}{38971--38988}.
\bibitem[{Dosovitskiy et~al.(2020)Dosovitskiy, Beyer, Kolesnikov, Weissenborn, Zhai, Unterthiner, Dehghani, Minderer, Heigold, Gelly et~al.}]{dosovitskiy2020image}
\bibinfo{author}{A.~Dosovitskiy}, \bibinfo{author}{L.~Beyer}, \bibinfo{author}{A.~Kolesnikov}, \bibinfo{author}{D.~Weissenborn}, \bibinfo{author}{X.~Zhai}, \bibinfo{author}{T.~Unterthiner}, \bibinfo{author}{M.~Dehghani}, \bibinfo{author}{M.~Minderer}, \bibinfo{author}{G.~Heigold}, \bibinfo{author}{S.~Gelly}, et~al.,
\newblock \bibinfo{title}{An image is worth 16x16 words: Transformers for image recognition at scale},
\newblock \bibinfo{journal}{arXiv preprint arXiv:2010.11929}  (\bibinfo{year}{2020}).
\bibitem[{Lu et~al.(2024)Lu, Lan, Zhang, Jiang, Wang, and Yuan}]{lu2024cricavpr}
\bibinfo{author}{F.~Lu}, \bibinfo{author}{X.~Lan}, \bibinfo{author}{L.~Zhang}, \bibinfo{author}{D.~Jiang}, \bibinfo{author}{Y.~Wang}, \bibinfo{author}{C.~Yuan},
\newblock \bibinfo{title}{Cricavpr: Cross-image correlation-aware representation learning for visual place recognition},
\newblock \bibinfo{journal}{arXiv preprint arXiv:2402.19231}  (\bibinfo{year}{2024}).
\bibitem[{Szegedy et~al.(2015)Szegedy, Liu, Jia, Sermanet, Reed, Anguelov, Erhan, Vanhoucke, and Rabinovich}]{szegedy2015going}
\bibinfo{author}{C.~Szegedy}, \bibinfo{author}{W.~Liu}, \bibinfo{author}{Y.~Jia}, \bibinfo{author}{P.~Sermanet}, \bibinfo{author}{S.~Reed}, \bibinfo{author}{D.~Anguelov}, \bibinfo{author}{D.~Erhan}, \bibinfo{author}{V.~Vanhoucke}, \bibinfo{author}{A.~Rabinovich},
\newblock \bibinfo{title}{Going deeper with convolutions},
\newblock in: \bibinfo{booktitle}{Proceedings of the IEEE conference on computer vision and pattern recognition}, \bibinfo{year}{2015}, pp. \bibinfo{pages}{1--9}.
\bibitem[{Ali-bey et~al.(2022)Ali-bey, Chaib-draa, and Gigu{\`e}re}]{ali2022gsv}
\bibinfo{author}{A.~Ali-bey}, \bibinfo{author}{B.~Chaib-draa}, \bibinfo{author}{P.~Gigu{\`e}re},
\newblock \bibinfo{title}{Gsv-cities: Toward appropriate supervised visual place recognition},
\newblock \bibinfo{journal}{Neurocomputing} \bibinfo{volume}{513} (\bibinfo{year}{2022}) \bibinfo{pages}{194--203}.
\bibitem[{Torii et~al.(2013)Torii, Sivic, Pajdla, and Okutomi}]{torii2013visual}
\bibinfo{author}{A.~Torii}, \bibinfo{author}{J.~Sivic}, \bibinfo{author}{T.~Pajdla}, \bibinfo{author}{M.~Okutomi},
\newblock \bibinfo{title}{Visual place recognition with repetitive structures},
\newblock in: \bibinfo{booktitle}{Proceedings of the IEEE conference on computer vision and pattern recognition}, \bibinfo{year}{2013}, pp. \bibinfo{pages}{883--890}.
\bibitem[{Warburg et~al.(2020)Warburg, Hauberg, Lopez-Antequera, Gargallo, Kuang, and Civera}]{warburg2020mapillary}
\bibinfo{author}{F.~Warburg}, \bibinfo{author}{S.~Hauberg}, \bibinfo{author}{M.~Lopez-Antequera}, \bibinfo{author}{P.~Gargallo}, \bibinfo{author}{Y.~Kuang}, \bibinfo{author}{J.~Civera},
\newblock \bibinfo{title}{Mapillary street-level sequences: A dataset for lifelong place recognition},
\newblock in: \bibinfo{booktitle}{Proceedings of the IEEE/CVF conference on computer vision and pattern recognition}, \bibinfo{year}{2020}, pp. \bibinfo{pages}{2626--2635}.
\bibitem[{Zaffar et~al.(2021)Zaffar, Garg, Milford, Kooij, Flynn, McDonald-Maier, and Ehsan}]{zaffar2021vpr}
\bibinfo{author}{M.~Zaffar}, \bibinfo{author}{S.~Garg}, \bibinfo{author}{M.~Milford}, \bibinfo{author}{J.~Kooij}, \bibinfo{author}{D.~Flynn}, \bibinfo{author}{K.~McDonald-Maier}, \bibinfo{author}{S.~Ehsan},
\newblock \bibinfo{title}{Vpr-bench: An open-source visual place recognition evaluation framework with quantifiable viewpoint and appearance change},
\newblock \bibinfo{journal}{International Journal of Computer Vision} \bibinfo{volume}{129} (\bibinfo{year}{2021}) \bibinfo{pages}{2136--2174}.
\bibitem[{Paszke et~al.(2019)Paszke, Gross, Massa, Lerer, Bradbury, Chanan, Killeen, Lin, Gimelshein, Antiga et~al.}]{paszke2019pytorch}
\bibinfo{author}{A.~Paszke}, \bibinfo{author}{S.~Gross}, \bibinfo{author}{F.~Massa}, \bibinfo{author}{A.~Lerer}, \bibinfo{author}{J.~Bradbury}, \bibinfo{author}{G.~Chanan}, \bibinfo{author}{T.~Killeen}, \bibinfo{author}{Z.~Lin}, \bibinfo{author}{N.~Gimelshein}, \bibinfo{author}{L.~Antiga}, et~al.,
\newblock \bibinfo{title}{Pytorch: An imperative style, high-performance deep learning library},
\newblock \bibinfo{journal}{Advances in neural information processing systems} \bibinfo{volume}{32} (\bibinfo{year}{2019}).
\bibitem[{Krizhevsky et~al.(2012)Krizhevsky, Sutskever, and Hinton}]{krizhevsky2012imagenet}
\bibinfo{author}{A.~Krizhevsky}, \bibinfo{author}{I.~Sutskever}, \bibinfo{author}{G.~E. Hinton},
\newblock \bibinfo{title}{Imagenet classification with deep convolutional neural networks},
\newblock \bibinfo{journal}{Advances in neural information processing systems} \bibinfo{volume}{25} (\bibinfo{year}{2012}).
\bibitem[{Wang et~al.(2019)Wang, Han, Huang, Dong, and Scott}]{wang2019multi}
\bibinfo{author}{X.~Wang}, \bibinfo{author}{X.~Han}, \bibinfo{author}{W.~Huang}, \bibinfo{author}{D.~Dong}, \bibinfo{author}{M.~R. Scott},
\newblock \bibinfo{title}{Multi-similarity loss with general pair weighting for deep metric learning},
\newblock in: \bibinfo{booktitle}{Proceedings of the IEEE/CVF conference on computer vision and pattern recognition}, \bibinfo{year}{2019}, pp. \bibinfo{pages}{5022--5030}.
\bibitem[{Loshchilov and Hutter(2017)}]{loshchilov2017decoupled}
\bibinfo{author}{I.~Loshchilov}, \bibinfo{author}{F.~Hutter},
\newblock \bibinfo{title}{Decoupled weight decay regularization},
\newblock \bibinfo{journal}{arXiv preprint arXiv:1711.05101}  (\bibinfo{year}{2017}).
\bibitem[{Berton et~al.(2023)Berton, Trivigno, Caputo, and Masone}]{berton2023eigenplaces}
\bibinfo{author}{G.~Berton}, \bibinfo{author}{G.~Trivigno}, \bibinfo{author}{B.~Caputo}, \bibinfo{author}{C.~Masone},
\newblock \bibinfo{title}{Eigenplaces: Training viewpoint robust models for visual place recognition},
\newblock in: \bibinfo{booktitle}{Proceedings of the IEEE/CVF International Conference on Computer Vision}, \bibinfo{year}{2023}, pp. \bibinfo{pages}{11080--11090}.

\end{thebibliography}

\end{document}